\documentclass[10pt,twocolumn,letterpaper]{article}
\usepackage[pagenumbers]{cvpr}
\usepackage{tikz}
\usepackage{pgfplots}
\pgfplotsset{compat=1.18}
\usetikzlibrary{
    shapes.geometric, arrows.meta, positioning,
    fit, calc, backgrounds, decorations.pathreplacing
}
\definecolor{colInput}{RGB}{236,240,241}
\definecolor{colPE}{RGB}{213,213,255}
\definecolor{colFine}{RGB}{255,218,170}
\definecolor{colCoarseOne}{RGB}{174,214,241}
\definecolor{colCoarseTwo}{RGB}{169,223,191}
\definecolor{colFuser}{RGB}{230,190,255}
\definecolor{colOutput}{RGB}{200,230,201}
\definecolor{colInset}{RGB}{255,253,220}
\definecolor{colArrow}{RGB}{80,80,80}
\definecolor{colBorder}{RGB}{110,110,110}
\pgfdeclarelayer{background}
\pgfdeclarelayer{middleground}
\pgfsetlayers{background,middleground,main}

\definecolor{cvprblue}{rgb}{0.21,0.49,0.74}
\usepackage[breaklinks,colorlinks,allcolors=cvprblue]{hyperref}
\hypersetup{
  pdftitle={PAM-Mamba - Efficient Dense Action Detection using Mamba},
  pdfauthor={Dhruvraj Singh Rawat; Faegheh Sardari; Armin Mustafa}
}
\title{PAM-Mamba - Efficient Dense Action Detection using Mamba}
\author{Dhruvraj Singh Rawat\quad
Faegheh Sardari\quad Armin Mustafa\\
Centre for Vision, Speech and Signal Processing (CVSSP)\\
University of Surrey, Guildford, UK}

\begin{document}
\maketitle
\begin{abstract}
Dense action detection requires precisely localising multiple overlapping actions in long untrimmed videos. Transformer-based methods achieve strong accuracy but their $O(T^2)$ self-attention cost limits scalability. We present \textbf{PAM-Mamba}, a linear-time $O(T)$ architecture built on Mamba~\cite{gu2024mamba} with a multi-branch temporal encoder. \textit{Gated Learnable Positional Encoding} (GLPE) controls positional signal injection via a scalar gate initialised to zero, starting training without added positional signal. A \textit{Hub-and-Spoke Funnel} extends the non-hierarchical multi-scale design of PAT~\cite{sardari2023position} with a direct residual from the original full-resolution input to each coarse branch, providing each coarse branch with a projection of the input features. \textit{Bi-TemBa} (Bidirectional Temporal Mamba) first extracts local motion cues with a shallow Temporal Convolution Module, then applies a cascaded bidirectional Mamba scan; the backward pass refines the forward-scanned representation from the opposite direction, giving each temporal position access to context from both directions in linear time. \textit{Uncertainty-Based Mixup} uses prediction entropy to guide the selection of mixup partners. Using the PAT evaluation implementation, we obtain \textbf{26.82} (Charades I3D), \textbf{34.59} (Charades CLIP), \textbf{46.56} (MultiTHUMOS I3D), and \textbf{46.92} mAP (MultiTHUMOS CLIP), with 26.1--27.9M parameters, 61--63\% fewer than PAT~\cite{sardari2023position}.

\end{abstract}
    
\section{Introduction}
\label{sec:intro}

Dense action detection asks a model to simultaneously answer \emph{what} actions are occurring, \emph{when} each begins and ends, and \emph{how many} overlap at any given moment, all within video sequences that can span tens of minutes~\cite{sigurdsson2016charades, yeung2018every}. The task demands frame-level precision across temporal scales ranging from sub-second gestures to multi-minute routines, with labels that co-occur densely and datasets that are overwhelmingly negative. Convolutional networks~\cite{lea2017temporal, farha2019ms} handle local patterns well but lack sufficient receptive field for long-range context. Transformers~\cite{dai2022ms, sardari2023position} address this through self-attention, but the $O(T^2)$ cost becomes prohibitive as sequence length grows.

State Space Models (SSMs), and Mamba~\cite{gu2024mamba} in particular, offer a compelling middle ground: linear-time $O(T)$ modelling with content-dependent state updates. Recent work has adapted Mamba to visual recognition~\cite{zhu2024visionmamba, ma2024mambavideo, islam2022long, lu2024videomambapro}, showing competitive accuracy at a fraction of the compute. Adapting Mamba to \emph{dense} action detection, however, raises four challenges that prior work has not jointly addressed.

\textbf{Positional encoding}: Mamba has implicit temporal order through its recurrent updates; the usefulness of an additional explicit positional signal must be established empirically. \textbf{Multi-scale reasoning}: multiple temporal resolutions offer complementary representations of brief and long activities; existing Mamba-based methods do not provide all temporal scales with direct access to the full-resolution input, so fine-grained boundary cues can be lost at coarser representations. \textbf{Non-causal context}: at an action onset, a causal model sees the current and preceding frames but cannot use subsequent frames to confirm that an action has started. Mamba's left-to-right recurrence makes boundary prediction at both onset and offset inherently one-sided. \textbf{Label imbalance}: dense multi-label datasets are dominated by negative frames, and the effect of mixing training examples depends on their labels and the batch configuration.

We present \textbf{PAM-Mamba} (\textbf{P}osition-\textbf{A}ware \textbf{M}amba), which combines three architectural components with an uncertainty-guided augmentation strategy:

\begin{itemize}

\item \textbf{Gated Learnable Positional Encoding (GLPE).} A scalar gate, initialised to zero, suppresses all positional signal at the start of training. As training proceeds, the gate learns how much explicit position information to inject, so the model incorporates it only to the extent that it helps.

\item \textbf{Hub-and-Spoke Funnel with Input Residuals.} Building on the non-hierarchical multi-scale design of PAT~\cite{sardari2023position}, we add a direct residual connection from the original full-resolution input to each coarse branch. Where PAT's coarse branches see only fine-processed features, our residual path supplies an independently projected input representation to each coarse scale.

\item \textbf{Bi-TemBa} (Bidirectional Temporal Mamba). A shallow Temporal Convolution Module (TCM) first captures local motion cues via Conv1d. MS-Temba~\cite{sinha2025ms} also uses temporal convolution; our block integrates local temporal processing within the proposed multi-branch encoder. A forward Mamba SSM then scans the TCM output; a backward SSM refines the forward representation from the opposite direction, giving each temporal position causal context from the left and anti-causal refinement from the right. This cascaded design differs from the independent parallel scans used in Vision Mamba and MS-Temba.

\item \textbf{Uncertainty-Based Mixup.} After an initial random-mixup phase, samples are ranked by Shannon entropy and the reversed entropy ranking supplies partner indices when the uncertainty gap exceeds the threshold. This extends standard random mixup~\cite{zhang2018mixup} with prediction-dependent partner selection (Sec.~\ref{subsec:uncertainty_mixup}).

\end{itemize}

\begin{figure*}[t]
  \centering
  \includegraphics[width=\linewidth]{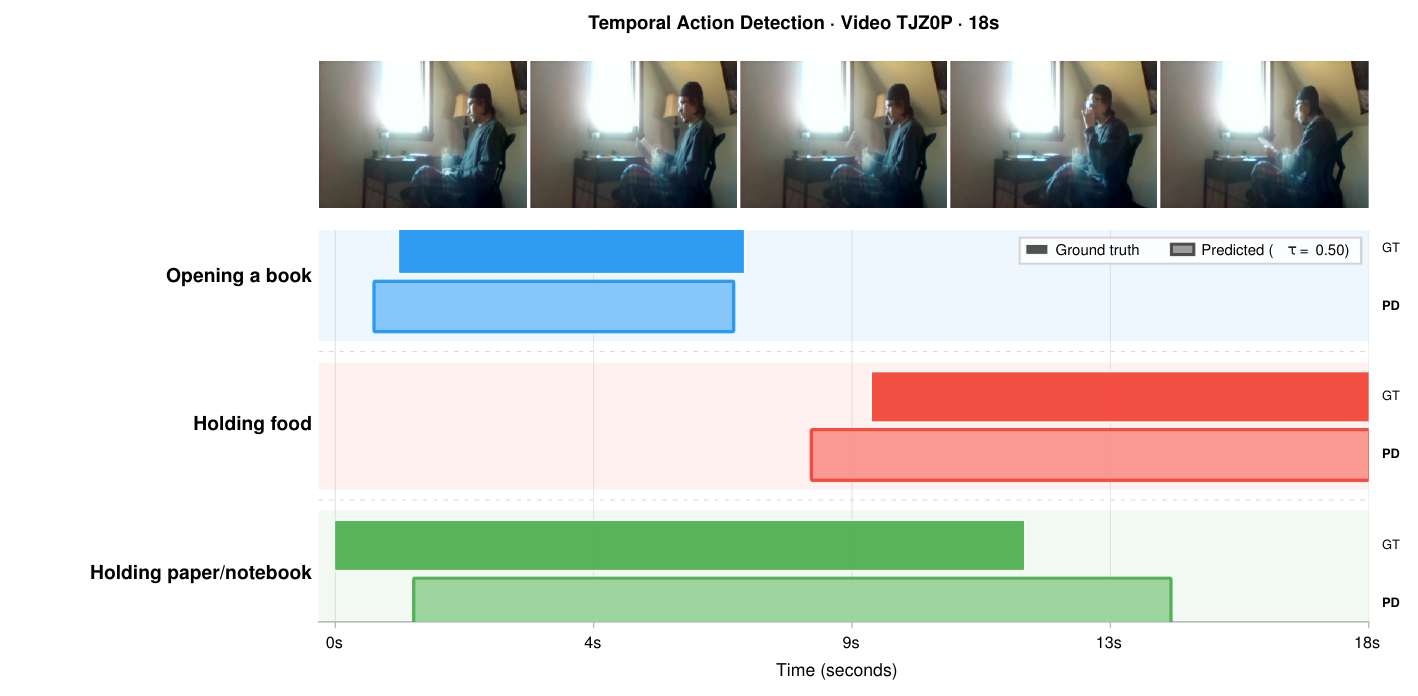}
  \caption{
    \textbf{PAM-Mamba qualitative result on Charades~\cite{sigurdsson2016charades}
    (CLIP~\cite{radford2021learning} features), video \texttt{TJZ0P}.}
    Each class pair shows the ground-truth segment (solid bar) and
    PAM-Mamba's predicted segment (translucent bar) at threshold $\tau\!=\!0.5$.
    Predicted intervals closely follow the annotated actions, with small onset and offset deviations.
    The concurrent actions \emph{Holding food} and \emph{Holding paper/notebook} are detected simultaneously over their shared interval.
  }
  \label{fig:temporal_strip}
\end{figure*}

Using the PAT evaluation implementation, we obtain \textbf{26.82}, \textbf{34.59}, \textbf{46.56}, and \textbf{46.92} mAP for I3D-Charades, CLIP-Charades, I3D-MultiTHUMOS, and CLIP-MultiTHUMOS respectively, with 61--63\% fewer parameters than PAT~\cite{sardari2023position} (Table~\ref{tab:main_results}). Figure~\ref{fig:temporal_strip} shows a qualitative example on a Charades video: PAM-Mamba detects \emph{Opening a book} and the concurrent actions \emph{Holding food} and \emph{Holding paper/notebook}, whose intervals overlap. These results motivate further study of state-space models for dense temporal prediction.

\section{Related Work}
\label{sec:related_work}

\subsection{Dense Action Detection}
Dense action detection (DAD) aims to localise and classify multiple concurrent human actions in long untrimmed videos. Early convolutional models, Temporal Convolutional Networks~\cite{lea2017temporal} and Multi-Stage TCNs~\cite{farha2019ms}, established strong baselines through dilated convolutions, but their fixed receptive fields limited long-range context modelling. Proposal-based approaches such as SSN~\cite{zhao2017temporal} and BMN~\cite{lin2019bmn} model candidate action intervals and their temporal boundaries. Their interval-level formulation differs from the frame-level multi-label prediction considered here.

\subsection{Transformer-Based Temporal Modelling}
The Transformer~\cite{vaswani2017attention} brought global temporal modelling to dense detection through self-attention. MS-TCT~\cite{dai2022ms} combined convolutional inductive bias with multi-head attention; PAT~\cite{sardari2023position} introduced non-hierarchical multi-scale fusion that preserves fine-grained features by generating coarse scales independently from a shared fine-resolution representation. Both achieve strong performance on Charades~\cite{sigurdsson2016charades} and MultiTHUMOS~\cite{yeung2018every}. The fundamental limitation is the $O(T^2)$ attention cost: Video Swin~\cite{liu2022video} uses local attention windows, whereas ViViT~\cite{arnab2021vivit} explores factorised spatial and temporal attention to reduce computation.

\subsection{State Space Models for Temporal Reasoning}
S4~\cite{gu2022efficiently} demonstrated efficient long-range dependency modelling; Mamba~\cite{gu2024mamba} extended this with content-dependent selective state updates. Vision Mamba~\cite{zhu2024visionmamba} applied bidirectional SSM scanning to 2D image patches for spatial recognition; VideoMamba~\cite{ma2024mambavideo} adapted this to spatiotemporal tokens. Islam and Bertasius~\cite{islam2022long} showed SSMs can classify long movie clips more efficiently than Transformers; VideoMambaPro~\cite{lu2024videomambapro} improved video representations through enhanced scanning strategies. Recurrent-Transformer hybrids like TRecViT~\cite{patraucean2024trecvit} explore replacing attention with recurrent modules for scalable video understanding. These approaches primarily target recognition rather than frame-level multi-label detection.

The most relevant SSM work for our setting is MS-Temba~\cite{sinha2025ms}, which applies Mamba to dense action detection. Each TemBa block combines temporal convolution with a dilated temporal SSM, and a scale-aware fuser aggregates block outputs. MS-Temba demonstrates that SSMs are viable for temporal action detection, but leaves open the questions of how to handle positional encoding in Mamba, how to provide each temporal scale with direct access to the full-resolution input, and how to address label imbalance in the augmentation strategy.

\subsection{Multi-Scale Architectures and Uncertainty-Aware Learning}
Feature Pyramid Networks~\cite{lin2017feature} established hierarchical top-down multi-scale fusion for 2D detection, merging coarse semantic features into finer spatial levels via lateral connections. TriDet~\cite{shi2023tridet} uses multi-scale temporal representations, while DualDETR~\cite{cheng2023dualdetr} combines instance-level and boundary-level queries. These methods address temporal localisation through different architectural choices. PAT~\cite{sardari2023position} addressed this with a non-hierarchical design: a fine-resolution module produces a single shared feature map, and multiple parallel coarse branches each subsample that map independently at different strides, avoiding cascaded information loss. However, all coarse branches in PAT derive exclusively from the fine-processed features, the original full-resolution input is not directly accessible at any coarse scale. On the learning side, uncertainty-aware training~\cite{kendall2017uncertainties} and mixup augmentation~\cite{zhang2018mixup} provide complementary approaches to training; we combine them by using prediction entropy to guide mixup partner selection for dense temporal detection.

\subsection{Summary}
Dense action detection has progressed from convolutional models~\cite{lea2017temporal, farha2019ms} to Transformer frameworks~\cite{sardari2023position, dai2022ms} and more recently to SSM-based architectures~\cite{gu2024mamba, sinha2025ms}. Transformers deliver strong accuracy but at quadratic cost; MS-Temba demonstrates SSMs can be competitive for temporal detection, yet leaves open key design questions around positional encoding, full-resolution input access at coarse scales, and training-time label imbalance. PAM-Mamba addresses these design questions while retaining linear-time temporal processing.

\begin{figure}[t]
  \centering
  \includegraphics[width=\linewidth]{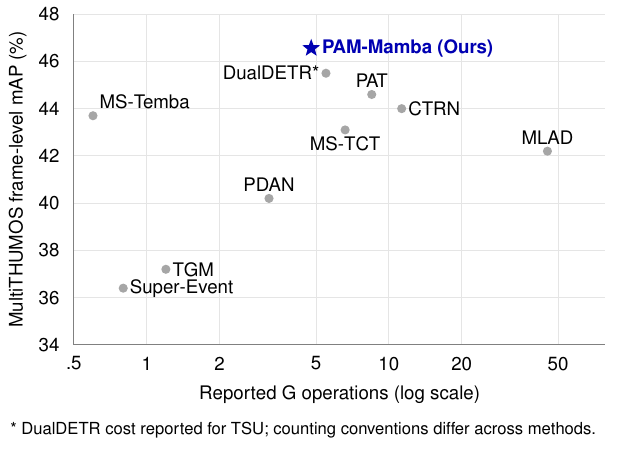}
  \caption{
  \textbf{Efficiency vs.\ accuracy on MultiTHUMOS (I3D features).}
  Per-frame mAP versus GFLOPs (log scale). PAM-Mamba achieves the highest mAP among the compared methods. Values for prior methods are taken from their publications; *DualDETR's GFLOPs are reported on TSU.
  }
  \label{fig:related_work}
\end{figure}

\section{Method}
\label{sec:method}

\begin{figure*}[t]
\centering
\definecolor{figInk}{RGB}{39,52,68}
\definecolor{figRule}{RGB}{112,130,148}
\definecolor{figBlue}{RGB}{235,242,249}
\definecolor{figGreen}{RGB}{235,245,240}
\definecolor{figPurple}{RGB}{241,237,248}
\definecolor{figWarm}{RGB}{249,244,232}
\resizebox{0.9\textwidth}{!}{%
\begin{tikzpicture}[
  font=\fontsize{9}{10.5}\selectfont\sffamily,
  text=figInk,
  block/.style={draw=figRule, fill=white, line width=0.6pt,
    rounded corners=1.5pt, align=center, text width=2.55cm,
    minimum height=0.84cm, inner xsep=3pt, inner ysep=3pt},
  group/.style={draw=figRule!65, line width=0.55pt, rounded corners=3pt},
  title/.style={font=\fontsize{9}{11}\selectfont\sffamily\bfseries,
    anchor=west, inner sep=0pt},
  tab/.style={title, fill=white, inner xsep=4pt, inner ysep=1.5pt},
  note/.style={font=\fontsize{9}{10.5}\selectfont, inner sep=1pt},
  arr/.style={draw=figInk,line width=0.75pt,
    -{Stealth[length=1.6mm,width=1.15mm]}},
  wire/.style={draw=figInk,line width=0.75pt},
  residual/.style={arr,dash pattern=on 2.5pt off 1.8pt},
  bridge/.style={preaction={draw=white,line width=3pt}},
  dot/.style={circle,fill=figInk,draw=none,inner sep=0pt,minimum size=3pt},
  sum/.style={circle,draw=figInk,fill=white,line width=0.65pt,
    inner sep=0pt,minimum size=0.38cm,font=\fontsize{10}{11}\selectfont},
]
\thickmuskip=4mu\medmuskip=3mu\thinmuskip=2mu
\begin{pgfonlayer}{background}
  \draw[group,fill=white] (0.25,-0.45) rectangle (12.75,-9.05);
  \draw[group,fill=figWarm] (0.4,-0.75) rectangle (12.6,-2.55);
  \draw[group,fill=figBlue] (0.4,-3.65) rectangle (5.9,-8.85);
  \draw[group,fill=figGreen] (7.1,-3.65) rectangle (12.6,-8.85);
  \draw[group,fill=figPurple] (0.4,-9.85) rectangle (12.6,-14.5);
  \draw[group,fill=figBlue!45] (13.6,-0.75) rectangle (17.5,-10.0);
  \draw[group,fill=white] (13.6,-10.4) rectangle (17.5,-12.3);
\end{pgfonlayer}

\node[block,text width=3.15cm,minimum height=1.15cm] (input) at (2.0,1.0)
  {\textbf{CLIP / I3D Features}\\
   $\mathbf{X}\in\mathbb{R}^{T\times D}$\\$D\in\{768,1024\}$};
\node[block,text width=3.85cm,minimum height=1.15cm,fill=figPurple] (gpe) at (6.5,1.0)
  {\textbf{Gated PE}\\
   $\mathbf{X}'=\mathbf{X}+\tanh(g)\cdot\mathrm{PE}$};
\node[block,text width=2.45cm,minimum height=1.15cm] (drop) at (10.7,1.0)
  {\textbf{Dropout}\\$p=0.5$};
\draw[arr] (input.east) -- (gpe.west);
\draw[arr] (gpe.east) -- (drop.west);

\node[tab] at (0.5,-0.45) {MambaEncoder};

\node[title,anchor=center] at (6.5,-1.05) {Fine Branch ($1\times$)};
\node[block] (fproj) at (3.1,-1.85)
  {Projection\\$D\to1024,\ s=1$};
\node[block,fill=figBlue] (ftemba) at (6.5,-1.85)
  {\textbf{Bi-TemBa}\\$d=1024$};
\node[block] (fnorm) at (9.9,-1.85)
  {LayerNorm\\$T\times1024$};
\draw[arr] (drop.south) -- (10.7,0.0) -- (3.1,0.0) -- (fproj.north);
\draw[arr] (fproj.east) -- (ftemba.west);
\draw[arr] (ftemba.east) -- (fnorm.west);
\node[dot] (fan) at (6.5,-3.2) {};
\draw[arr] (fnorm.south) -- (9.9,-2.85) -- (6.5,-2.85) -- (fan.north);

\node[title] at (0.65,-3.95) {Coarse-1 ($2\times\downarrow$)};
\node[title,anchor=east] at (12.35,-3.95) {Coarse-2 ($4\times\downarrow$)};
\node[block,text width=2.42cm,minimum height=1.35cm] (c1res) at (1.75,-5.05)
  {Projection\\\mbox{\fontsize{8.5}{10}\selectfont$D\to768,\ s=2$}\\\textit{from} $\mathbf{X}'$};
\node[block,text width=2.42cm,minimum height=1.35cm] (c1fine) at (4.55,-5.05)
  {Projection\\\mbox{\fontsize{8.5}{10}\selectfont$1024\to768,\ s=2$}\\\textit{from Fine}};
\node[sum] (c1add) at (3.15,-6.25) {$+$};
\node[block,fill=figBlue] (c1temba) at (3.15,-7.0)
  {\textbf{Bi-TemBa}\\$d=768$};
\node[block] (c1norm) at (3.15,-8.2)
  {LayerNorm\\$T/2\times768$};
\draw[arr] (c1res.south) -- (1.75,-6.25) -- (c1add.west);
\draw[arr] (c1fine.south) -- (4.55,-6.25) -- (c1add.east);
\draw[arr] (c1add.south) -- (c1temba.north);
\draw[arr] (c1temba.south) -- (c1norm.north);

\node[block,text width=2.42cm,minimum height=1.35cm] (c2fine) at (8.45,-5.05)
  {Projection\\\mbox{\fontsize{8.5}{10}\selectfont$1024\to512,\ s=4$}\\\textit{from Fine}};
\node[block,text width=2.42cm,minimum height=1.35cm] (c2res) at (11.25,-5.05)
  {Projection\\\mbox{\fontsize{8.5}{10}\selectfont$D\to512,\ s=4$}\\\textit{from} $\mathbf{X}'$};
\node[sum] (c2add) at (9.85,-6.25) {$+$};
\node[block,fill=figBlue] (c2temba) at (9.85,-7.0)
  {\textbf{Bi-TemBa}\\$d=512$};
\node[block] (c2norm) at (9.85,-8.2)
  {LayerNorm\\$T/4\times512$};
\draw[arr] (c2fine.south) -- (8.45,-6.25) -- (c2add.west);
\draw[arr] (c2res.south) -- (11.25,-6.25) -- (c2add.east);
\draw[arr] (c2add.south) -- (c2temba.north);
\draw[arr] (c2temba.south) -- (c2norm.north);

\draw[arr] (fan.center) -- (4.55,-3.2) -- (c1fine.north);
\draw[arr] (fan.center) -- (8.45,-3.2) -- (c2fine.north);

\draw[residual] (3.1,0.0) -- (0.0,0.0) -- (0.0,-5.05) -- (c1res.west);
\draw[residual] (10.7,0.0) -- (13.0,0.0) -- (13.0,-5.05) -- (c2res.east);
\node[dot] at (3.1,0.0) {};
\node[dot] at (10.7,0.0) {};
\node[note,anchor=south west] at (0.05,0.05) {$\mathbf{X}'$};
\node[note,anchor=south east] at (12.95,0.05) {$\mathbf{X}'$};

\node[title] at (0.65,-10.15) {SimpleFuser};
\node[block,minimum height=1.1cm] (projF) at (2.2,-11.1)
  {Linear\\$1024\to1024$\\\textit{from Fine}};
\node[block,minimum height=1.1cm] (projC1) at (6.8,-11.1)
  {Linear\\$768\to1024$};
\node[block,minimum height=1.1cm] (projC2) at (10.8,-11.1)
  {Linear\\$512\to1024$};
\draw[arr] (c1norm.south) -- (3.15,-9.6) -- (6.8,-9.6) -- (projC1.north);
\draw[arr] (c2norm.south) -- (9.85,-9.6) -- (10.8,-9.6) -- (projC2.north);
\draw[wire] (fan.center) -- (6.5,-9.35);
\draw[wire,bridge] (6.5,-9.35) -- (0.1,-9.35);
\draw[arr] (0.1,-9.35) -- (0.1,-11.1) -- (projF.west);

\node[block,text width=4.2cm] (fuse) at (8.8,-12.55)
  {Upsample \& weighted sum\\$T\times1024$};
\draw[arr] (projC1.south) -- (6.8,-11.85) -- (7.8,-11.85)
  -- ([xshift=-1cm]fuse.north);
\draw[arr] (projC2.south) -- (10.8,-11.85) -- (9.8,-11.85)
  -- ([xshift=1cm]fuse.north);
\node[block] (headF) at (2.2,-13.8)
  {Classifier$_{\mathrm{fine}}$\\$T\times C$};
\node[block] (headC) at (8.8,-13.8)
  {Classifier$_{\mathrm{coarse}}$\\$T\times C$};
\draw[arr] (projF.south) -- (headF.north);
\draw[arr] (fuse.south) -- (headC.north);

\node[block,text width=6.6cm,minimum height=0.62cm,fill=figGreen] (combine) at (6.5,-15.25)
  {$\mathbf{Z}=(1-\lambda)\,\mathbf{Z}_{\mathrm{fine}}+\lambda\,\mathbf{Z}_{\mathrm{coarse}},\quad\lambda=0.75$};
\draw[arr] (headF.south) -- (2.2,-14.75) -- (5.5,-14.75)
  -- ([xshift=-1cm]combine.north);
\draw[arr] (headC.south) -- (8.8,-14.75) -- (7.5,-14.75)
  -- ([xshift=1cm]combine.north);
\node[block,text width=4.2cm,minimum height=0.62cm,fill=figGreen] (pred) at (6.5,-16.25)
  {Probabilities $\hat{\mathbf{Y}}=\sigma(\mathbf{Z})$};
\draw[arr] (combine.south) -- (pred.north);

\node[title] at (13.85,-1.07) {Bi-TemBa Block};
\begin{scope}[yshift=0.75cm]
\node[block,text width=2.3cm,minimum height=0.55cm] (bx) at (15.55,-2.5)
  {$\mathbf{x}\in\mathbb{R}^{T\times d}$};
\node[block,text width=2.3cm,minimum height=0.55cm] (btcm) at (15.55,-3.4) {TCM};
\node[block,text width=2.3cm,minimum height=0.55cm] (bfwd) at (15.55,-4.3) {LN $\to$ Mamba$_{\mathrm{f}}$};
\node[block,text width=2.3cm,minimum height=0.55cm] (bflipin) at (15.55,-5.2) {Reverse time};
\node[block,text width=2.3cm,minimum height=0.55cm] (bbwd) at (15.55,-6.1) {LN $\to$ Mamba$_{\mathrm{b}}$};
\node[block,text width=2.3cm,minimum height=0.55cm] (bflipout) at (15.55,-7.0) {Reverse time};
\node[sum] (bsum) at (15.55,-7.8) {$+$};
\node[block,text width=2.3cm,minimum height=0.55cm] (bnorm) at (15.55,-8.6) {LayerNorm};
\node[sum] (badd) at (15.55,-9.4) {$+$};
\node[block,text width=2.3cm,minimum height=0.55cm] (bout) at (15.55,-10.2) {Output $\mathbf{x}+\mathbf{H}'$};
\draw[arr] (bx.south) -- (btcm.north);
\draw[arr] (btcm.south) -- (bfwd.north);
\draw[arr] (bfwd.south) -- (bflipin.north);
\draw[arr] (bflipin.south) -- (bbwd.north);
\draw[arr] (bbwd.south) -- (bflipout.north);
\draw[arr] (bflipout.south) -- (bsum.north);
\draw[arr] (bfwd.west) -- (13.95,-4.3) -- (13.95,-7.8) -- (bsum.west);
\draw[arr] (bsum.south) -- (bnorm.north);
\draw[arr] (bnorm.south) -- (badd.north);
\draw[residual] (bx.east) -- (17.2,-2.5) -- (17.2,-9.4) -- (badd.east);
\draw[arr] (badd.south) -- (bout.north);
\end{scope}
\node[title] at (13.85,-10.7) {Legend};
\draw[arr] (13.9,-11.3) -- (14.7,-11.3);
\node[note,anchor=west,font=\fontsize{8.5}{10}\selectfont\sffamily] at (14.9,-11.3) {Data flow};
\draw[residual] (13.9,-11.9) -- (14.7,-11.9);
\node[note,anchor=west,font=\fontsize{8.5}{10}\selectfont\sffamily] at (14.9,-11.9) {Input residual};
\end{tikzpicture}%
}
\caption{\textbf{PAM-Mamba architecture overview.}
Pre-extracted I3D or CLIP features receive gated learnable positional encoding (including dropout 0.1) and input dropout 0.5 before entering the three-branch encoder. The full-resolution fine branch supplies a shared representation to two independently downsampled coarse branches. Each coarse branch also receives a strided projection of the encoder input. Projection blocks include LayerNorm.
The Bi-TemBa inset follows the implementation: TCM is Conv1d, LayerNorm and GELU; the forward scan feeds the backward scan through temporal reversal. The backward output is reversed back, summed with the forward output and normalised, then added to the original block input. Both Mamba scans use pre-normalisation.
SimpleFuser projects branch outputs to 1024 dimensions, linearly upsamples coarse outputs and combines them using learned softmax weights. Separate fine and coarse classifiers produce logits $\mathbf{Z}_{\mathrm{fine}}$ and $\mathbf{Z}_{\mathrm{coarse}}$, blended with the reported 0.25/0.75 weights; a sigmoid gives per-class probabilities.}
\label{fig:architecture}
\end{figure*}
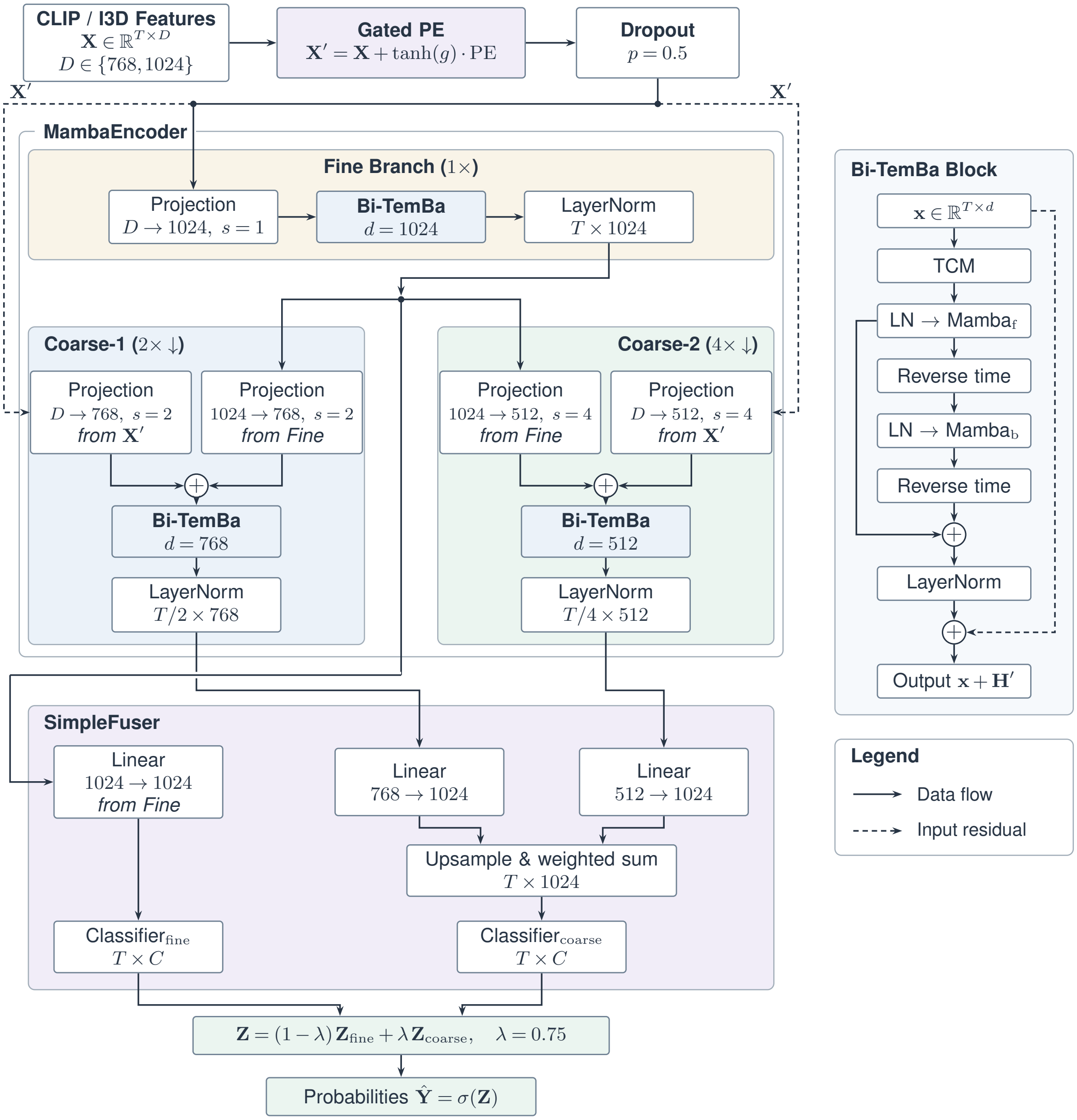

We propose a multi-branch Mamba encoder for dense action detection. Actions span vastly different durations, from half-second gestures to multi-minute activities, and multiple temporal scales can provide complementary context. Mamba's causal recurrence also means boundary frames have no access to future context, which is critical for onset and offset prediction. \textbf{PAM-Mamba} addresses these limitations through four components whose interactions are shown in Figure~\ref{fig:architecture}.

\subsection{Overall Architecture}

PAM-Mamba takes as input a sequence of $T$ pre-extracted feature vectors $\mathbf{x} \in \mathbb{R}^{T \times d}$, one per temporal feature position, from a pretrained backbone (I3D~\cite{carreira2017quo}, $d\!=\!1024$; CLIP~\cite{radford2021learning}, $d\!=\!768$). The pipeline has four stages. The $O(T)$ complexity refers to temporal processing of pre-extracted features with fixed channel widths, state dimensions and block counts; it does not include visual feature extraction.

\noindent\textbf{(1) Gated Learnable Positional Encoding (GLPE).} A scalar gate, initialised to zero, controls how much explicit position information is added to $\mathbf{x}$, yielding $\mathbf{x}' \in \mathbb{R}^{T \times d}$ (Sec.~\ref{subsec:gated_pe}).

\noindent\textbf{(2) Hub-and-Spoke Funnel.} $\mathbf{x}'$ is processed by a full-resolution fine branch to produce $x_\mathrm{fine} \in \mathbb{R}^{T \times C_1}$. Two coarse branches each receive both downsampled fine features and a direct residual projection of $\mathbf{x}'$, producing representations at $T/2$ and $T/4$ (Sec.~\ref{subsec:residual_funnel}).

\noindent\textbf{(3) Bi-TemBa blocks.} Each branch is processed by a Bi-TemBa block that pairs a local Temporal Convolution Module with cascaded bidirectional Mamba scanning, enabling non-causal temporal modelling at each scale in linear time (Sec.~\ref{subsec:bidirectional_mamba}).

\noindent\textbf{(4) SimpleFuser and prediction heads.} Each branch output, fine ($C_1$-d at resolution $T$) and the two coarse branches ($C_2$-d at $T/2$, $C_3$-d at $T/4$), is first projected to a common $D_f$-d space via a per-branch linear layer. Coarse outputs are then upsampled to resolution $T$ by linear interpolation. A softmax-normalised scalar weight per coarse branch produces a single weighted-sum coarse feature. Distinct linear heads score the coarse and fine features; at inference their logits are combined at a fixed ratio (see Sec.~\ref{subsec:setup}) and a sigmoid gives per-position probabilities over all $C$ classes. Asymmetric Loss~\cite{ridnik2021asymmetric} is applied throughout to handle the severe label imbalance of dense multi-label detection.

\subsection{Gated Learnable Positional Encoding}
\label{subsec:gated_pe}

Positional encodings play a different role in Mamba than in Transformers. Self-attention without positional information is permutation-equivariant, so Transformers cannot distinguish token order without explicit position signals~\cite{vaswani2017attention}. Mamba, by contrast, processes sequences recurrently: each hidden state $h_t$ is conditioned on all prior tokens in their original order, giving the model implicit positional awareness without any additional encoding. We therefore treat the strength of an additional explicit positional signal as a learnable quantity rather than assuming that a fixed signal is beneficial.

A second consideration is the relation between clip positions and physical time: temporal cropping and changes in sampling rate can alter this relation. Learned position embeddings can adapt to the training distribution, but do not explicitly encode timestamps.

We propose \textbf{Gated Learnable Positional Encoding (GLPE)}. A learnable embedding matrix $\mathbf{P} \in \mathbb{R}^{T_{\max} \times d}$ is gated by a scalar parameter $g$, initialised to $0$:
\begin{equation}
  \mathbf{x}' = \mathbf{x} + \tanh(g)\cdot\mathbf{P}_{1:T},
  \label{eq:glpe}
\end{equation}
where $\mathbf{P}_{1:T}$ selects the first $T$ rows of $\mathbf{P}$. The implementation sets $T_{\max}=5000$ and initialises the table with a truncated normal distribution with standard deviation 0.02. GLPE applies dropout with probability 0.1, followed by a separate input dropout with probability 0.5. At initialisation, $\tanh(0)=0$, the gate is fully closed and training begins with no positional injection. As $g$ is updated by gradient descent, $\tanh(g)$ can take positive or negative values, allowing the strength and sign of positional injection to adapt; it need not approach either saturation limit. The learnable matrix $\mathbf{P}$ further adapts to the temporal distribution of the training data, unlike fixed encodings. The ablations show dataset-dependent effects of GLPE (Sec.~\ref{subsec:ablations}).

\subsection{Hub-and-Spoke Funnel with Input Residuals}
\label{subsec:residual_funnel}

Multi-scale temporal representations are essential for detecting actions across widely different durations. PAT~\cite{sardari2023position} established a key design principle for this task: rather than cascading scales hierarchically (as in FPN~\cite{lin2017feature}), a non-hierarchical \emph{hub} produces a shared fine-resolution feature map, and each coarse branch independently subsamples it at a different stride. This avoids the information loss of sequential downsampling chains, where errors at one scale propagate to all finer scales.

PAT's coarse branches, however, derive exclusively from the fine-processed features, the original full-resolution input is not directly accessible at any coarse scale. Our \textbf{Hub-and-Spoke Funnel} preserves the non-hierarchical topology and adds a \textbf{direct residual from the original input} to each coarse branch. The fine branch processes the GLPE-encoded input:
\begin{equation}
  x_\mathrm{fine} = \mathrm{TemBa}\!\left(\phi^{\mathrm{in}}(\mathbf{x}')\right),
\end{equation}
where $\phi^{\mathrm{in}}$ is a stride-1 Conv1d projection to embedding dimension $C_1$. Each coarse branch $s \in \{1, \ldots, S\}$ receives two independent inputs:
\begin{equation}
  x_s = \mathrm{TemBa}\!\left(\phi_s^{h}(x_\mathrm{fine}) + \phi_s^{r}(\mathbf{x}')\right),
  \label{eq:hubspoke}
\end{equation}
where $\phi_s^{h}$ downsamples the fine branch to resolution $T/2^s$ (the \emph{hub path}), and $\phi_s^{r}$ directly projects $\mathbf{x}'$ to the same resolution (the \emph{residual path}). Both are lightweight strided Conv1d blocks with LayerNorm. The coarse branches remain independent of each other. Each projection uses kernel size 3 and padding 1, with strides 1, 2 and 4 for the fine and two coarse branches. A separate LayerNorm follows each Bi-TemBa block; this branch-output normalisation is implicit in the branch equations.

The residual path $\phi_s^{r}(\mathbf{x}')$ ensures that each coarse branch retains direct access to the original input features, independent of any abstraction introduced by the fine branch. Its benefit is most pronounced on the long videos of MultiTHUMOS (Sec.~\ref{subsec:ablations}).

\subsection{Bi-TemBa: Bidirectional Temporal Mamba Block}
\label{subsec:bidirectional_mamba}

Boundary localisation can draw on two complementary signals at each temporal position: \emph{local motion cues} (gesture dynamics, short-range transitions) and \emph{long-range context} (activity duration, co-occurrence patterns). Standard causal Mamba~\cite{gu2024mamba} processes sequences left-to-right only, which creates a fundamental asymmetry at action boundaries. At an onset frame, the first frame where an action is active, the model has access to the current frame and preceding frames, but not subsequent frames; the strongest evidence that the action has started is in the frames immediately \emph{after}, where the activity is underway. A causal model must commit to a boundary decision before seeing that evidence. The same holds at the offset: the frames after an action ends confirm its cessation, but they are invisible to a left-to-right model. Non-causal context therefore provides additional evidence for offline boundary prediction.

Vision Mamba~\cite{zhu2024visionmamba} introduced bidirectional SSM scanning for 2D image recognition by running independent forward and backward Mamba layers on raw spatial patch tokens. MS-Temba~\cite{sinha2025ms} adapted bidirectional SSMs to temporal action detection, combining a temporal convolution module with dilated temporal SSMs to capture multiple temporal scales. \textbf{Bi-TemBa} (Bidirectional Temporal Mamba) is a block for dense temporal detection that combines the following operations.

First, Bi-TemBa uses a \textbf{Temporal Convolution Module (TCM)}: a shallow Conv1d with LayerNorm and GELU activation. Unlike a linear layer, Conv1d operates over a local temporal window, explicitly encoding short-range motion patterns, gesture dynamics, transient pose changes, that are discriminative near action boundaries. These locally-enriched features are then passed to the SSM for long-range propagation. Temporal convolution is also used by MS-Temba~\cite{sinha2025ms}; its presence alone is not a distinction from that method.

Second, rather than running the two scanning directions independently, Bi-TemBa uses a \textbf{cascaded bidirectional scan}: the forward SSM processes the TCM output first, and the backward SSM then takes the reversed forward representation as input, rather than the reversed TCM output directly. The backward pass therefore refines and completes the temporal context already established by the forward pass; every frame's final encoding carries both causal context from the left and anti-causal refinement informed by that context from the right.

Third, a \textbf{block-level residual} connects the block input directly to its output, bypassing both the TCM and the SSM. This provides a direct path for the block input alongside the convolutional and state-space operations. The full computation is:
\begin{align}
\mathbf{H}_{\mathrm{tcm}} &= \mathrm{GELU}(\mathrm{LN}(\mathrm{Conv1d}(\mathbf{x}))), \label{eq:tcm}\\
\mathbf{H}_{\mathrm{fwd}} &= \mathrm{Mamba}_{\mathrm{fwd}}(\mathrm{LN}_{f}(\mathbf{H}_{\mathrm{tcm}})), \label{eq:bitemba_fwd}\\
\mathbf{H}_{\mathrm{bwd}} &= \overline{\mathrm{Mamba}_{\mathrm{bwd}}(\mathrm{LN}_{b}(\overline{\mathbf{H}_{\mathrm{fwd}}}))},\\
\mathbf{H}' &= \mathrm{LN}_{o}(\mathbf{H}_{\mathrm{fwd}} + \mathbf{H}_{\mathrm{bwd}}), \label{eq:bitemba}\\
\mathbf{x} &\leftarrow \mathbf{x} + \mathbf{H}', \label{eq:residual}
\end{align}
where $\overline{\cdot}$ denotes temporal reversal and $\mathrm{LN}_{f}$, $\mathrm{LN}_{b}$ and $\mathrm{LN}_{o}$ are separate LayerNorm operations. The implementation uses one forward/backward pair per branch. TCM uses a kernel of size 3, stride 1 and padding 1. Forward and backward outputs are summed rather than gated, avoiding the additional learnable parameters that a gated fusion would require.

\subsection{Uncertainty-Based Mixup Augmentation}
\label{subsec:uncertainty_mixup}

Standard mixup~\cite{zhang2018mixup} interpolates randomly chosen sample pairs. In dense multi-label detection, random pairing does not explicitly prioritise examples with uncertain predictions. \textbf{Uncertainty-Based Mixup} addresses this with a curriculum: training begins with standard random mixup and transitions to entropy-guided pairing after a fixed warm-up period. Entropy is used as a proxy for uncertainty, rather than a direct measurement of sample difficulty or calibration.

Concretely, the implementation applies random mixup during the initial $E_0$ epochs. Thereafter, it computes sigmoid predictions in evaluation mode without gradients. The uncertainty of clip $i$ is binary entropy averaged over classes and temporal positions:
\begin{align}
\mathcal{H}(x_i) &= \frac{1}{TC}\sum_{t=1}^{T}\sum_{c=1}^{C} h(p_{itc}),\\
h(p) &= -p\log p-(1-p)\log(1-p).
\end{align}
Let $q$ contain the sample indices sorted by decreasing entropy. If the batch entropy range exceeds $\delta$, the code assigns partners using the reversed index list, $\pi(i)=q_{B+1-i}$, while retaining the original order of the first operand. Otherwise it uses a random permutation. Inputs and frame-level targets are mixed with the same coefficient:
\begin{align}
\tilde{x}_i &= \lambda x_i+(1-\lambda)x_{\pi(i)},\\
\tilde{y}_i &= \lambda y_i+(1-\lambda)y_{\pi(i)},\quad
\lambda\sim\mathrm{Beta}(\alpha,\alpha).
\end{align}
Pairing therefore depends on the batch: with batch size one, mixup reduces to the identity, and larger batches offer more candidate partners.

\section{Experiments}
\label{sec:experiments}

\begin{figure*}[p]
  \centering
  \includegraphics[width=\linewidth]{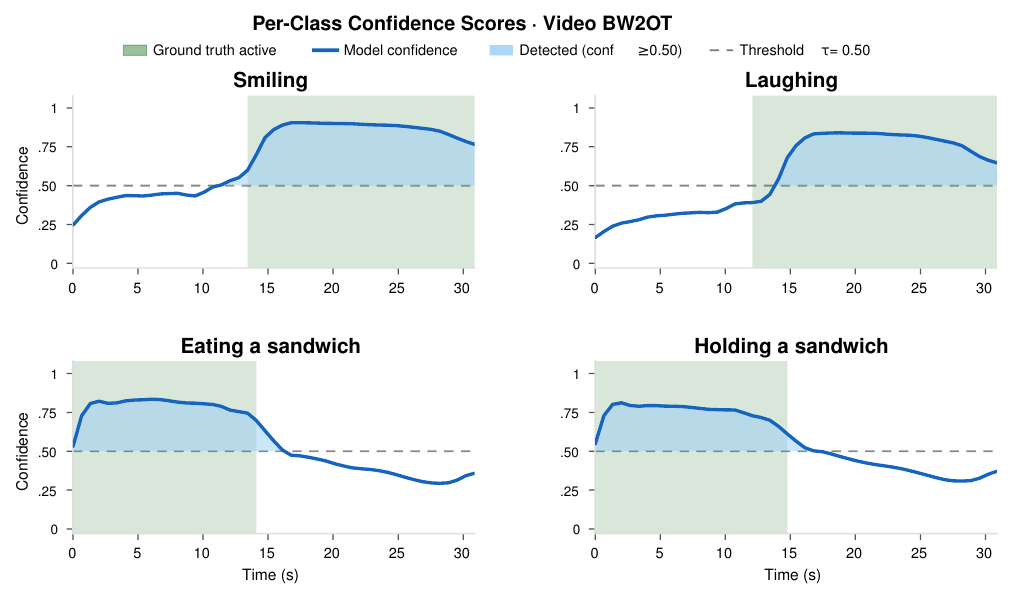}
  \caption{
    \textbf{PAM-Mamba per-class confidence on Charades (CLIP features), video \texttt{BW2OT}.}
    Each panel shows the sigmoid output for one action class over time.
    The shaded green region marks the ground-truth active interval; the blue
    shaded area shows positions where the model fires above threshold $\tau\!=\!0.5$.
    The confidence profiles closely track the annotated intervals.
    The two concurrent actions \emph{Eating a sandwich} and \emph{Holding a sandwich}
    (bottom row) are detected simultaneously over their shared interval.
  }
  \label{fig:confidence_curves}
\end{figure*}

We evaluate \textbf{PAM-Mamba} on the standard dense action detection benchmarks \textbf{Charades}~\cite{sigurdsson2016charades} and \textbf{MultiTHUMOS}~\cite{yeung2018every}, reporting frame-level mean Average Precision (mAP) using the PAT evaluation implementation and comparing against prior work~\cite{sardari2023position, dai2022ms, sinha2025ms}. Figure~\ref{fig:confidence_curves} shows per-class confidence on a test video: confidence rises within the annotated action intervals, and the bottom two panels show concurrent detections.

\subsection{Experimental Setup}
\label{subsec:setup}

Similar to PAT~\cite{sardari2023position}, during both training and inference PAM-Mamba uses a fixed number of $T=256$ input tokens, where each token is a pre-extracted I3D~\cite{carreira2017quo} or CLIP~\cite{radford2021learning} feature vector. Labels are assigned to tokens from the annotated start and end times, and per-frame metrics are computed over these tokens. For training, we randomly sample a clip containing $T$ consecutive tokens from a video sequence and zero-pad shorter sequences. At inference, we use the PAT evaluation code, which takes the first $T$ tokens of each test sequence. We use one forward/backward Mamba pair per branch, $d_{\mathrm{state}}=18$, branch widths 1024/768/512 and a 1024-dimensional fusion space. Coarse/fine logits and losses are weighted 0.75/0.25. Python, NumPy, PyTorch and CUDA seeds are set to 42; deterministic cuDNN behaviour is enabled and cuDNN benchmarking disabled.

For \textbf{Charades}, both feature settings use AdamW~\cite{loshchilov2017decoupled}, weight decay 0.25, batch size 2, up to 50 epochs and StepLR with step 12 and $\gamma=0.1$. I3D uses learning rate $2\times10^{-4}$ and patience 15; CLIP uses $10^{-4}$ and patience 5. Drop-path rate is configured as 0.3. Mixup uses $\alpha=0.2$, threshold 0.25 and transition epoch 5.

For \textbf{MultiTHUMOS}, both settings use batch size 1, up to 100 epochs, patience 20 and configured drop-path rate 0.4. I3D uses learning rate $10^{-5}$, weight decay 0.30 and StepLR step 15; CLIP uses $10^{-4}$, weight decay 0.35 and step 40, with $\gamma=0.1$ for both. Mixup is enabled in the configuration but is an identity operation at batch size 1 (Sec.~\ref{subsec:uncertainty_mixup}). The configured $(\alpha,\delta,E_0)$ values are $(0.25,0.25,10)$ for I3D and $(0.2,0.3,10)$ for CLIP.

Asymmetric Loss~\cite{ridnik2021asymmetric} uses $\gamma_{\mathrm{pos}}=1$ and clipping 0.1. The negative exponent is 3 for Charades, 5 for MultiTHUMOS I3D and 8 for MultiTHUMOS CLIP. Gradients are clipped to norm 1.0.

\noindent\textbf{Stochastic Weight Averaging (SWA).}
We enable SWA~\cite{izmailov2018averaging} from epoch 10 on Charades with learning rate $5\times10^{-5}$, and epoch 30 on MultiTHUMOS with learning rate $10^{-5}$. The implementation updates averaged weights once per epoch after this point and uses SWALR in place of StepLR. Base and averaged models are evaluated separately; the logged best score is the maximum across these evaluations.

\noindent\textbf{Evaluation implementation.}
We use the PAT evaluation code~\cite{sardari2023position} unchanged, including its per-frame AP computation, probability masking and averaging over nonzero class APs, and evaluate on the standard test split. Every Charades test sequence fits within $T$ (at most 227 I3D or 109 CLIP tokens), so Charades predictions cover full videos. For MultiTHUMOS, 166 of 211 I3D and 177 of 211 CLIP test sequences are longer than $T$.

\subsection{Datasets and Metrics}
\label{subsec:datasets}

\noindent\textbf{Charades}~\cite{sigurdsson2016charades} contains 9{,}848 untrimmed videos across 157 activity classes with strong label co-occurrence. The official split contains 7{,}985 training and 1{,}863 testing videos. Following PAT, test videos without action annotations are excluded, leaving 1{,}814 test videos.

\noindent\textbf{MultiTHUMOS}~\cite{yeung2018every} extends THUMOS14 to 413 videos with dense frame-level annotations for 65 classes, with an average of 1.5 labels per frame and 10.5 distinct action classes per video. These statistics distinguish frame-level overlap from the number of classes occurring anywhere in a video.

\noindent\textbf{Metric.} Following PAT~\cite{sardari2023position}, we report per-frame mAP (\%). AP is computed for each class by ranking frame predictions and averaging precision at positive labels. We retain the PAT implementation's convention of averaging nonzero class APs.

\subsection{Comparison with State of the Art}
\label{subsec:main_results}

Table~\ref{tab:main_results} lists PAM-Mamba and prior results. The reported I3D model size is 27.9M parameters, compared with 70.7M for PAT~\cite{sardari2023position}; MS-Temba~\cite{sinha2025ms} is smaller at 10.1M. Our CLIP configuration has a reported size of 26.1M. The architecture is shared across feature settings, with input dimensions of 1024 for I3D and 768 for CLIP.

\noindent\textbf{Computation.} We measure GFLOPs with \texttt{thop} on an input of $T=256$ tokens: 4.78G for I3D and 4.48G for CLIP. Values for prior methods are taken from their publications; DualDETR's GFLOPs are reported on TSU~\cite{cheng2023dualdetr}, and the CLIP values for MS-TCT and MS-Temba are from~\cite{sinha2025ms}.

\begin{table*}[t]
  \centering
  \caption{
  Comparison on Charades and MultiTHUMOS using I3D and CLIP features.
  Values are per-frame mAP (\%); best results in bold. $^{a}$GFLOPs reported on TSU. $^{b}$Obtained by training the official PAT code on the same CLIP features.
  }
  \label{tab:main_results}
  \resizebox{\textwidth}{!}{
  \begin{tabular}{lccccc}
  \toprule
  \textbf{Method} & \textbf{Features} & \textbf{Params (M)} & \textbf{GFLOPs} & \textbf{Charades} & \textbf{MultiTHUMOS} \\
  \midrule
  \multicolumn{6}{c}{\textit{I3D Features}} \\
  \midrule
  Super-Event~\cite{piergiovanni2018super} & I3D & 26 & 0.8 & 18.6 & 36.4 \\
  TGM~\cite{piergiovanni2019temporal} & I3D & 2 & 1.2 & 20.6 & 37.2 \\
  PDAN~\cite{wang2021pdan} & I3D & 6 & 3.2 & 23.7 & 40.2 \\
  MLAD~\cite{tirupattur2021modeling} & I3D & 21 & 45.2 & 18.4 & 42.2 \\
  PointTAD~\cite{bai2020pointtad} & I3D & – & – & 21.0 & 39.8 \\
  HAAN~\cite{wang2023haan} & I3D & – & – & 25.2 & 41.7 \\
  CTRN~\cite{huang2020cttr} & I3D & 11 & 11.3 & 25.3 & 44.0 \\
  MS-TCT~\cite{dai2022ms} & I3D & 87.5 & 6.6 & 25.4 & 43.1 \\
  DualDETR~\cite{cheng2023dualdetr} & I3D & 21 & $5.5^{a}$ & 23.2 & 45.5 \\
  MS-Temba~\cite{sinha2025ms} & I3D & 10.1 & 0.6 & 25.3 & 43.7 \\
  PAT~\cite{sardari2023position} & I3D & 70.7 & 8.5 & 26.5 & 44.6 \\
  \midrule
  \textbf{PAM-Mamba (Ours)} & I3D & 27.9 & 4.78 & \textbf{26.82} & \textbf{46.56} \\
  \midrule
  \multicolumn{6}{c}{\textit{CLIP Features (TTM uses ViViT)}} \\
  \midrule
  TTM~\cite{ryoo2023token} & ViViT & 89 & – & 26.3 & – \\
  AAN~\cite{park2022aan} & CLIP & – & – & 32.0 & – \\
  MS-TCT~\cite{dai2022ms,sinha2025ms} & CLIP & 87.5 & 6.6 & 31.9 & 41.6 \\
  MS-Temba~\cite{sinha2025ms} & CLIP & 10.1 & 0.6 & 32.3 & 41.9 \\
  PAT$^{b}$~\cite{sardari2023position} & CLIP & 70.7 & – & 33.32 & 46.78 \\
  \midrule
  \textbf{PAM-Mamba (Ours)} & CLIP & 26.1 & 4.48 & \textbf{34.59} & \textbf{46.92} \\
  \bottomrule
  \end{tabular}
  }
\end{table*}

\noindent\textbf{Action-Conditional Metrics.}
We report action-conditional metrics introduced by \citet{tirupattur2021modeling} and used by PAT~\cite{sardari2023position}. The evaluator computes precision, recall, F1-score (F1AC) and mean AP (mAP\textsubscript{AC}) at $\tau\in\{0,20,40\}$ prediction time steps, using probability threshold 0.5. $\tau$ is measured in tokens. At $\tau=0$ the metric measures co-occurrence; positive windows consider preceding actions that are no longer active at the current step.

On Charades with I3D features, PAM-Mamba achieves F1AC of 27.9/31.8/31.6 and mAP\textsubscript{AC} of 30.1/36.4/35.4 at $\tau=0/20/40$. PAT reports F1AC of 27.2/29.6/29.4 and mAP\textsubscript{AC} of 32.0/37.8/36.7, and MS-TCT F1AC of 19.5/22.1/22.1~\cite{sardari2023position}. PAM-Mamba obtains the highest F1AC at every $\tau$, while PAT retains higher mAP\textsubscript{AC}; note that these prior results use both RGB and optical-flow inputs.

\begin{figure*}[p]
  \centering
  \includegraphics[width=\linewidth]{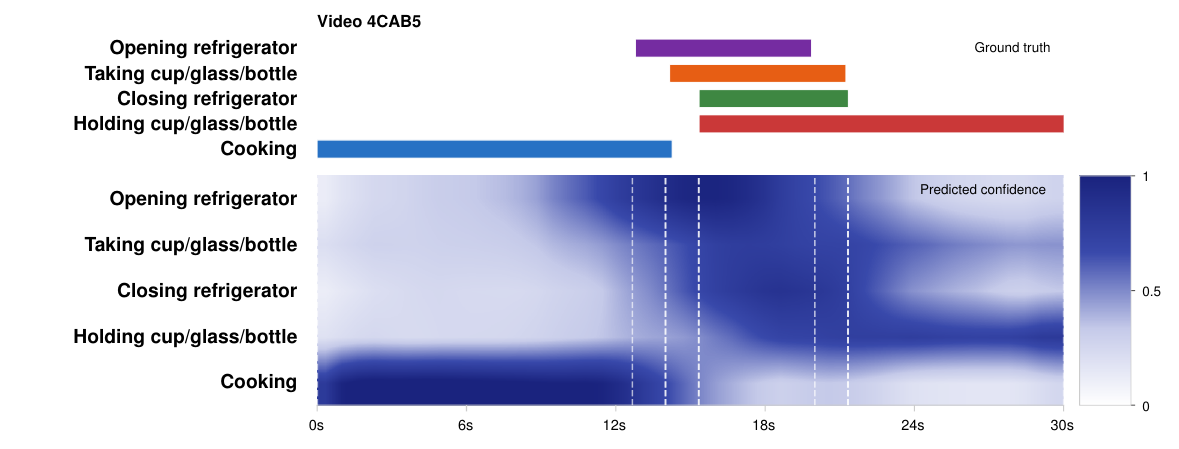}
  \caption{
    \textbf{Per-class confidence heatmap on Charades~\cite{sigurdsson2016charades}
    (CLIP~\cite{radford2021learning} features), video \texttt{4CAB5}.}
    \emph{Top:} ground-truth action segments colour-coded by class.
    \emph{Bottom:} PAM-Mamba's predicted confidence for each class over time
    (white $=$ low, deep indigo $=$ high).
    White dashed lines mark ground-truth onset and offset boundaries.
    Five of the seven annotated classes are shown. Confidence concentrates within the ground-truth intervals, including the tightly interleaved refrigerator and cup/glass/bottle actions.
  }
  \label{fig:multiscale_heatmap}
\end{figure*}

Figure~\ref{fig:related_work} plots accuracy against computation on MultiTHUMOS. Figure~\ref{fig:multiscale_heatmap} shows confidence over time for five co-occurring classes; high-confidence responses concentrate within the annotated intervals, with softer transitions at action boundaries.

\subsection{Ablation Studies}
\label{subsec:ablations}

\begin{figure}[t]
  \centering
  \includegraphics[width=\linewidth]{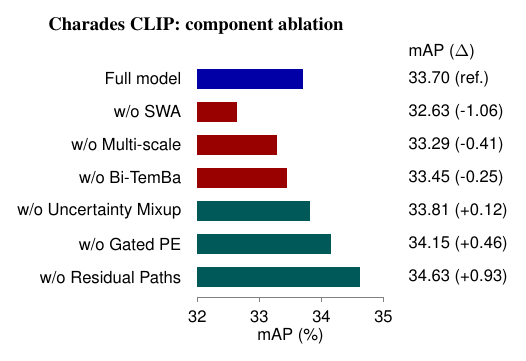}
  \caption{
    \textbf{Ablation study on Charades (CLIP features); see Table~\ref{tab:ablation} for MultiTHUMOS.}
    Labels show mAP and the difference from the full model (33.70).
  }
  \label{fig:ablation}
\end{figure}

Table~\ref{tab:ablation} and Figure~\ref{fig:ablation} report component ablations with CLIP features. Each variant removes one component from the full model, and differences are measured against the full-model run of the same ablation series (33.70 on Charades and 46.13 on MultiTHUMOS).

\begin{table}[t]
  \centering
  \caption{
    \textbf{Component ablation on Charades and MultiTHUMOS (CLIP features).}
    Each variant removes one component; $\Delta$ is relative to the full model of the same ablation series.
  }
  \label{tab:ablation}
  \setlength{\tabcolsep}{2.8pt}
  \begin{tabular}{lcccc}
    \toprule
    & \multicolumn{2}{c}{\textbf{Charades}} & \multicolumn{2}{c}{\textbf{MultiTHUMOS}} \\
    \cmidrule(lr){2-3}\cmidrule(lr){4-5}
    \textbf{Variant} & mAP & $\Delta$ & mAP & $\Delta$ \\
    \midrule
    Full Model & 33.70 & -- & 46.13 & -- \\
    w/o SWA & 32.63 & $-1.06$ & 44.72 & $-1.40$ \\
    w/o Multi-Scale & 33.29 & $-0.41$ & 45.67 & $-0.46$ \\
    w/o Bi-TemBa & 33.45 & $-0.25$ & 44.87 & $-1.25$ \\
    w/o Uncertainty Mixup & 33.81 & $+0.12$ & 45.25 & $-0.87$ \\
    w/o Gated PE & 34.15 & $+0.46$ & 45.37 & $-0.75$ \\
    w/o Residual Paths & 34.63 & $+0.93$ & 44.99 & $-1.13$ \\
    \bottomrule
  \end{tabular}
\end{table}

\noindent\textbf{SWA, multi-scale processing and bidirectionality.}
Removing SWA reduces mAP by 1.06 on Charades and 1.40 on MultiTHUMOS. Removing multi-scale processing reduces it by 0.41 and 0.46, respectively; removing bidirectionality reduces it by 0.25 and 1.25. These components therefore contribute consistently on both datasets, with SWA providing the largest gain.

\noindent\textbf{Mixup, positional encoding and input residuals.}
On MultiTHUMOS, removing GLPE or the input residual paths reduces mAP by 0.75 and 1.13, respectively. On Charades, whose videos are short and fit entirely within a single $T$-token window, removing them increases mAP by 0.46 and 0.93. Explicit positional information and direct access to full-resolution input thus matter most for long sequences. Mixup has little effect on Charades ($+0.12$ when removed). Because MultiTHUMOS is trained with batch size one, for which mixup reduces to the identity, its $-0.87$ difference reflects run-to-run variation rather than an effect of mixup.

\section{Conclusion}
\label{sec:conclusion}

We presented \textbf{PAM-Mamba}, a linear-time architecture for dense action detection with three architectural components and an uncertainty-guided augmentation strategy: Gated Learnable Positional Encoding, which lets the model control how much explicit position information it incorporates; a Hub-and-Spoke Funnel that extends PAT's non-hierarchical multi-scale design~\cite{sardari2023position} with direct input residuals to each coarse branch, providing an independent input projection at each coarse scale; Bi-TemBa, a block that pairs local convolution with a cascaded bidirectional Mamba scan for non-causal temporal reasoning; and Uncertainty-Based Mixup, which uses prediction entropy to select mixup partners when the batch contains multiple examples. Ablations show that SWA, multi-scale processing and bidirectional scanning improve both benchmarks, while GLPE and input residuals are most beneficial on the long videos of MultiTHUMOS.

With 26.1--27.9M parameters, PAM-Mamba achieves 26.82, 34.59, 46.56 and 46.92 mAP on Charades and MultiTHUMOS with I3D and CLIP features, showing that state-space models are a strong and efficient alternative to self-attention for dense action detection.

\noindent\textbf{Limitations.} The model operates on pre-extracted features and fixed-length clip segments. As visible in Figure~\ref{fig:temporal_strip}, precise boundary localisation remains a challenge: predicted segments overlap the annotated intervals but have onset and offset discrepancies, suggesting that additional boundary-sensitive losses could further improve temporal precision. Future work includes end-to-end training from raw video, extension to multimodal and streaming settings~\cite{girdhar2023omnivore}, and evaluation on egocentric benchmarks where dense multi-label detection is equally relevant.

{
    \small
    \bibliographystyle{ieeenat_fullname}
    \bibliography{main}
}

\end{document}